\documentclass[10pt,twocolumn,letterpaper]{article}

\usepackage{iccv}
\usepackage{times}
\usepackage{epsfig}
\usepackage{graphicx}
\usepackage{amsmath}
\usepackage{amssymb}
\usepackage{subcaption}
\usepackage[numbers,sort]{natbib} 
\usepackage{bbm}
\usepackage{algcompatible}
\usepackage{algpseudocode}
\usepackage{algorithm}
\algnewcommand\algorithmicforeach{\textbf{for each}}
\algdef{S}[FOR]{ForEach}[1]{\algorithmicforeach\ #1\ \algorithmicdo}

\algnewcommand\INPUT{\item[\textbf{Input:}]}%
\algnewcommand\OUTPUT{\item[\textbf{Output:}]}%
\usepackage[normalem]{ulem}

\usepackage{array}
\usepackage{tabularray}

\usepackage[para]{threeparttable} 
\usepackage[export]{adjustbox}
\usepackage{tabulary,color}
\usepackage{tabularx}
\usepackage{multirow}
\usepackage{booktabs, siunitx, dcolumn}
\usepackage[accsupp]{axessibility}  
\usepackage{authblk}
\usepackage{pdfpages}

\usepackage[pagebackref=true,breaklinks=true,letterpaper=true,colorlinks,bookmarks=false]{hyperref}

\iccvfinalcopy 

\ificcvfinal\fi

\begin{document}

\title{Camera-Driven Representation Learning for \\ Unsupervised Domain Adaptive Person Re-identification \vspace{-3mm}}
\author{%
Geon Lee$^{1}$ \hspace{1mm} 
Sanghoon Lee$^{1}$ \hspace{1mm} 
Dohyung Kim$^{1}$ \hspace{1mm}       
Younghoon Shin$^{2}$ \hspace{1mm} 
Yongsang Yoon $^{2}$  \hspace{1mm} 
{Bumsub Ham$^{1}$\thanks{Corresponding author}}  \\
$^{1}$Yonsei University \hspace{3mm} 
$^{2}$Robotics Lab, Hyundai Motor Company  \\ 
\vspace{1mm}
{\url{https://cvlab.yonsei.ac.kr/projects/CaCL}}
}

\maketitle

\ificcvfinal\thispagestyle{empty}\fi

\begin{abstract}
   We present a novel unsupervised domain adaption method for person re-identification~(reID) that generalizes a model trained on a labeled source domain to an unlabeled target domain. We introduce a camera-driven curriculum learning~(CaCL) framework that leverages camera labels of person images to transfer knowledge from source to target domains progressively. To this end, we divide target domain dataset into multiple subsets based on the camera labels, and initially train our model with a single subset~(i.e., images captured by a single camera). We then gradually exploit more subsets for training, according to a curriculum sequence obtained with a camera-driven scheduling rule. The scheduler considers maximum mean discrepancies~(MMD) between each subset and the source domain dataset, such that the subset closer to the source domain is exploited earlier within the curriculum. For each curriculum sequence, we generate pseudo labels of person images in a target domain to train a reID model in a supervised way. We have observed that the pseudo labels are highly biased toward cameras, suggesting that person images obtained from the same camera are likely to have the same pseudo labels, even for different IDs. To address the camera bias problem, we also introduce a camera-diversity~(CD) loss encouraging person images of the same pseudo label, but captured across various cameras, to involve more for discriminative feature learning, providing person representations robust to inter-camera variations. Experimental results on standard benchmarks, including real-to-real and synthetic-to-real scenarios, demonstrate the effectiveness of our framework.
\end{abstract}
\vspace{-2mm}

\section{Introduction}
\label{sec:intro}

\begin{figure}[t]
   \begin{center}
      \includegraphics[width=.95\linewidth]{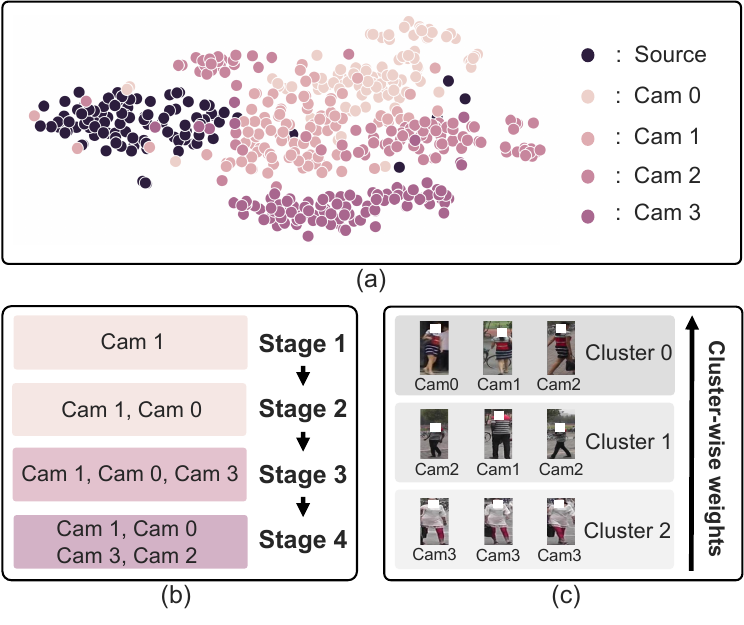}
   \end{center}
    \vspace{-5mm}  
   \caption{We visualize in (a) a t-SNE plot for features extracted from person images in Market1501~\cite{zheng2015scalable} and MSMT17~\cite{wei2018person}, using a reID model trained on MSMT17, where MSMT17 and Market1501 are source and target domains, respectively. The samples from different cameras in the target domain are distinguished by different colors. The model trained on a single domain offers features that are highly biased towards camera labels of person images for other domains. We propose to establish a camera-driven curriculum, as shown in (b), and initially train our model using images captured by a single camera, then gradually exploit more images captured using multiple cameras. To further alleviate the camera bias issue, we compute cluster-wise weights, as in (c), to encourage clusters containing images obtained from various cameras to involve more during the adaptation process.}
   \label{fig:teaser}
\end{figure}

The objective of person re-identification~(reID) is to retrieve person images of the same ID as a query person across non-overlapping cameras~\cite{ye2021deep, zheng2016person}. Current reID approaches mainly adopt a supervised learning paradigm by exploiting person ID labels, and focus on learning discriminative person representations in a single domain. However, reID models trained on a specific domain typically fail to generalize to other domains~\cite{wei2018person, deng2018image}, thus limiting the applicability in real-world scenarios. To address this issue, recent works~\cite{dai2021idm, zheng2021group, zheng2021exploiting, wei2018person} exploit unsupervised domain adaptation techniques, transferring knowledge learned from a source domain to re-identify persons in a target one, where ID labels for the source domain are provided only~\cite{ma2013domain}. This enables performing reID on the target domain without additional annotations, which is typically time-consuming and labor-intensive to obtain~\cite{zheng2015scalable, li2014deepreid, ristani2016performance}. Unsupervised domain adaptive (UDA) reID is challenging due to the following reasons. Transferring knowledge from one domain to another is difficult due to the distribution gap between camera topologies for different domains~\cite{ristani2016performance, zheng2015scalable, wei2018person}. Moreover, it is difficult to learn discriminative person representations for the target domain without ID labels, due to the large intra-class variations, particularly between person images captured by different cameras.

In recent years, most UDA reID methods~\cite{he2022secret, li2022reliability, dai2021idm, zheng2021online, isobe2021towards, zheng2021group, ge2020mutual, zhao2020unsupervised, zheng2021exploiting, ge2020self} exploit pseudo ID labels for target images to mitigate the discrepancies between source and target domains. To generate pseudo labels for the target domain, these methods first extract features from target images, using a reID model pre-trained on the source domain, and apply a clustering algorithm~(\eg, DBSCAN~\cite{ester1996density}) on the features. They then assign the same ID label to the images which belong to the same cluster, facilitating training with target images in a supervised manner. While the UDA reID methods have allowed significant advances for UDA reID, they mainly have two limitations. First, current approaches still focus on transferring knowledge from source to target in a domain-level. Namely, they attempt to adapt a model trained on a source domain to the target one \emph{at once}, by regarding target images as a whole. This is not effective for transferring knowledge for UDA reID, since source and target domains have different camera topologies. Second, pseudo labels for the target domain are highly biased towards camera labels of images~(Fig.~\ref{fig:teaser}(a)). That is, person images captured by the same camera are likely to be assigned to the same pseudo ID label, even for the persons with different IDs. Directly training a reID model with such labels rather hinders discriminative feature learning~\cite{dai2021idm,zheng2021exploiting,zheng2021group}, particularly for the person images of the same ID but captured by different cameras.

In this paper, we present a novel framework for UDA reID that performs a \emph{progressive} adaptation exploiting camera labels of person images. We conjecture that domain adaptation in a domain-level regime might be suboptimal, especially in the context of reID, since the distribution of a camera topology is highly unique for each domain. In order to consider an abrupt change on the camera topology from source to target domains, we propose a camera-driven curriculum learning~(CaCL) leveraging camera labels of person images, facilitating a progressive adaptation~(Fig.~\ref{fig:teaser}(b)). To implement this idea, we first decompose a target domain dataset into multiple subsets w.r.t the camera labels. Starting from a single subset~(\ie, images obtained from a single camera), we gradually add subsets to train our model, according to a curriculum sequence obtained by a camera-driven scheduling rule. The scheduler considers maximum mean discrepancies~(MMD)~\cite{gretton2006kernel} between each subset and the source domain dataset, such that a closer subset w.r.t the source dataset is exploited earlier within the curriculum. We also introduce a camera-diversity (CD) loss that encourages the clusters having person images obtained from various cameras to involve more for discriminative feature learning~(Fig.~\ref{fig:teaser}(c)). It further incorporates a selective scheme for training that discards trivial clusters, only consisting of person images taken from the same camera. A model trained with CD loss is able to offer person representations more robust to inter-camera variations, compared to conventional cross-entropy~\cite{zheng2017discriminatively} and triplet~\cite{hermans2017defense} losses, even when training with pseudo labels biased to camera labels. Together with CaCL and the CD loss, we achieve a new state of the art on standard UDA reID benchmarks, including real-to-real~(\eg, Market1501~\cite{zheng2015scalable}-to-MSMT17~\cite{wei2018person} and MSMT17-to-Market1501) and synthetic-to-real~(\eg, PersonX~\cite{sun2019dissecting}-to-Market1501 and Unreal~\cite{zhang2021unrealperson}-to-MSMT17) scenarios, and demonstrate the effectiveness of our approach with extensive experimental results and ablative analyses.
 
 Our main contributions can be summarized as follows: (1) We introduce a novel curriculum learning framework for UDA reID that leverages camera labels of person images. To the best of our knowledge, this is the first to incorporate a curriculum learning scheme for UDA reID. We also present the camera-driven scheduler that determines the curriculum sequence for multiple subsets in a target domain. (2) We present the CD loss to learn discriminative person representations, particularly robust to inter-camera variations, even when training with pseudo labels biased to camera labels. (3) We set a new state of the art on standard benchmarks for UDA reID, including real-to-real and synthetic-to-real scenarios, and demonstrate the effectiveness of our framework.

\section{Related work}
\label{sec:relatedworks}
\noindent \textbf{UDA reID.} There are many attempts for UDA reID to handle the domain gap between source and target, without ID labels for the target domain, which can be categorized into two groups. The first line of works~\cite{deng2018image, wei2018person} use generative models to translate person images of the source domain into the target one. Taking source images as input, target-stylized person images are generated using generative adversarial networks~(GANs)~\cite{goodfellow2014generative} for image translation~\cite{zhu2017unpaired}, with identity-preserving techniques~\cite{deng2018image}. The generated images are then used to train a reID model on the target domain in a supervised manner. These approaches to exploiting generative models typically involve many heuristics~\cite{salimans2016improved}, and require a lot of parameters, due to the unstable training of GANs~\cite{kodali2017convergence}. Another line of works~\cite{he2022secret, li2022reliability, dai2021idm, zheng2021online, isobe2021towards, zheng2021group, ge2020mutual, zhao2020unsupervised, zheng2021exploiting, ge2020self} transfer knowledge from source to target domains using a self-training scheme~\cite{fan2018unsupervised}. These methods pre-train a reID model on the source domain with ID labels, and exploit the model to extract person representations from target images. They apply a clustering algorithm on the representations, and person images within the same cluster are assigned the same pseudo ID label. As the quality of pseudo labels largely influences the reID performance on the target domain, the works of~\cite{zheng2021exploiting, zhao2020unsupervised, zheng2021online, ge2020mutual} attempt to refine the pseudo labels, \eg, by leveraging multiple reID models and measuring prediction consistencies~\cite{zhao2020unsupervised, zheng2021exploiting}. We have observed that the person representations for target images, obtained using the source-pretrained model, offer clustering results that are highly biased towards camera labels. In this context, camera labels of person images can provide complementary information to alleviate this problem, which has not been considered previously. Moreover, all the aforementioned approaches do not consider the large discrepancies between camera topologies for different domains. Therefore, they handle the domain gap in a domain-level regime, considering the target domain as a whole. Instead of mitigating the domain gap at once, a recent approach~\cite{dai2021idm} proposes to generate person representations of intermediate domains to perform adaptation gradually, by mixing cross-domain features~\cite{verma2019manifold}. This approach, however, still exploits all target images jointly during the adaptation process. In contrast to this, we start training with a subset of person images in the target domain, and gradually expand to multiple subsets, through a camera-driven curriculum for a progressive adaptation.

\noindent \textbf{ReID with auxiliary supervision.} Person reID methods focus on extracting discriminative person representations to match person of the same ID effectively, while differentiating persons of different IDs. Since it is challenging to handle large intra-class variations~(\eg, background clutter, viewpoint, and pose variations) with ID labels alone, many works exploit auxiliary supervisory signals for reID. Examples of the auxiliary signals include human pose~\cite{cao2017realtime, ge2018fd, zheng2019pose}, semantic parsing~\cite{kalayeh2018human}, and attribute labels~\cite{lin2019improving, schumann2017person}. These provide additional cues for, \eg, a part-to-part matching~\cite{kalayeh2018human} or a feature disentanglement~\cite{cao2017realtime}, enhancing the discriminative power of person representations. The auxiliary labels are expensive to obtain, as they use additional networks~\cite{cao2017realtime, wei2016convolutional, li2020self} trained with task-specific datasets~\cite{gong2017look, andriluka20142d} or require labor-intensive annotations~\cite{deng2014pedestrian, li2016richly}. Camera labels of input images, on the other hand, provide an efficient alternative, since they can easily be accessed from the metadata of images~\cite{zhong2018camera}. There are attempts to leverage camera labels of input images during training, to learn person representations robust to inter-camera variations, which is particularly important for reID that performs person matching in a cross-camera setting. For example, the work of~\cite{zhuang2020rethinking} computes camera-specific feature statistics to mitigate the distribution gap between different cameras. In the context of UDA reID, the work of~\cite{qi2019novel} proposes to use a discriminator for camera labels within an adversarial learning framework. The work of~\cite{zhong2018camera} generates multiple images of the same person in the style of different cameras, and uses the synthesized person images for UDA reID. In contrast to the UDA reID approaches to exploiting camera labels~\cite{zhuang2020rethinking, qi2019novel, zhong2018camera,li2022reliability}, we leverage the labels to establish a curriculum sequence and mitigate the bias for pseudo labels towards camera labels.

\vspace{1mm}
\noindent \textbf{Curriculum learning.} The seminal work of~\cite{bengio2009curriculum} introduces a curriculum learning strategy that trains a model using easy examples in early stages and with hard ones in later stages. Since then, the curriculum learning paradigm~\cite{soviany2022curriculum, wang2021survey} is adopted for various applications, including object detection~\cite{shrivastava2016training}, semantic segmentation~\cite{sakaridis2019guided}, and image synthesis~\cite{karras2017progressive}. The main difference between these methods lies in how they define easy and hard examples. Previous methods typically define specific criteria to establish curriculum sequences, \eg, by measuring distances to object boundaries for semantic segmentation~\cite{li2017not}, computing loss values of training samples~\cite{han2018co}, and employing a module for estimating difficulties of samples~\cite{yang2020curriculum}. More specific for UDA segmentation, a domain discriminator~\cite{sakaridis2019guided} and a pixel-wise label distribution~\cite{zhang2017curriculum} are used to define easy and hard samples. On the other hand, we incorporate camera labels of person images to facilitate a curriculum learning paradigm for UDA reID. We conjecture that camera topologies play a significant role in learning discriminative features, particularly for the task of person reID. To our knowledge, no previous approaches have incorporated camera topologies to set a curriculum.
 
\begin{figure*}[t]
   \begin{center}
   \includegraphics[width=.9\linewidth]{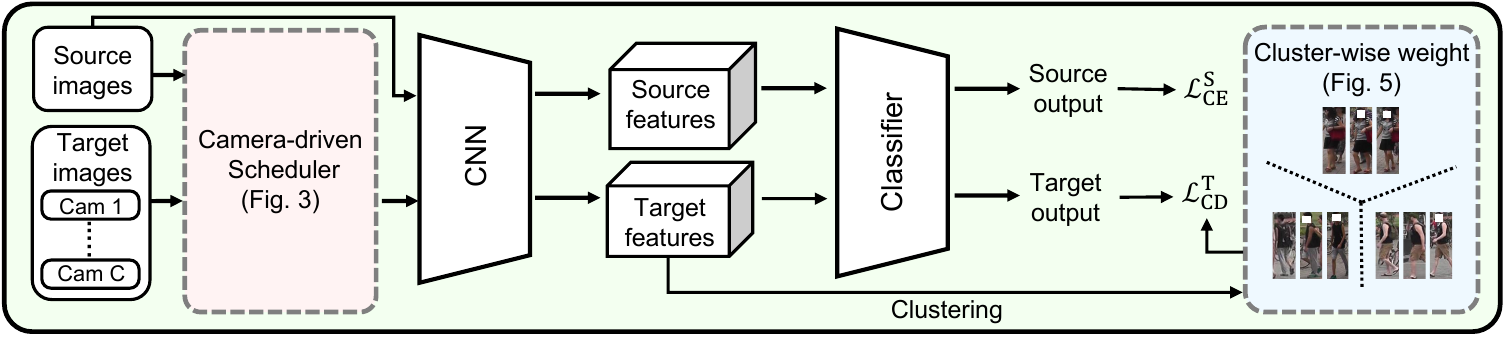}
   \end{center}
  	\vspace{-5mm}
   \caption{An overview of our framework. We divide target images into multiple subsets based on camera labels. The camera-driven scheduler takes the subsets of the target domain, along with source images as inputs, to establish a curriculum sequence. We train our model progressively with CD loss for a target domain~$\mathcal{L}_\textrm{CD}^{\textrm{T}}$, along with the cross-entropy term~\cite{zheng2017discriminatively} for a source domain~$\mathcal{L}_\textrm{CE}^{\textrm{S}}$. See text for more details.}
   \label{fig:overview}
\end{figure*}

\begin{figure}[t]
   \begin{center}
   \includegraphics[width=.9\linewidth]{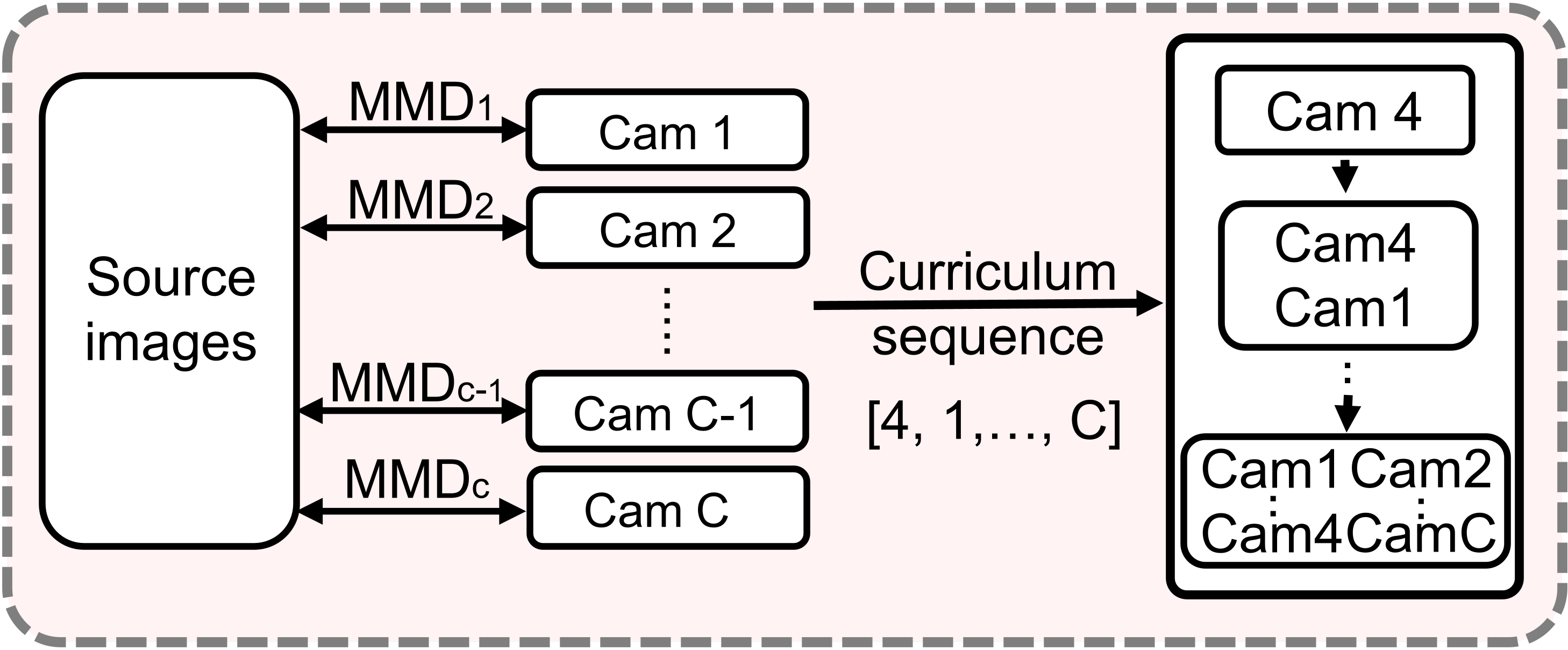}
   \end{center}
   \vspace{-5mm}
   \caption{An illumination of a camera-driven scheduler. We compute pairwise MMDs between source domain and all target subsets, establishing a curriculum sequence. We initially train our model with a single subset and gradually expand a training set by adding subsets in the sequence.}
   \label{fig:curriculum}
\end{figure}

\vspace{-2mm}

\section{Method}

\subsection{Overview}
\label{subsec:overview}

We provide in Fig.~\ref{fig:overview} an overview of our framework for UDA reID. We first divide a target dataset into multiple subsets by leveraging camera labels of person images. The camera-driven scheduler takes the target subsets, along with source images as inputs, to establish a curriculum sequence. Within each curriculum sequence, we adopt a self-training scheme~\cite{fan2018unsupervised} and alternate between clustering and fine-tuning. Specifically, we apply a clustering algorithm on person features extracted from target images to generate pseudo ID labels, and further fine-tune our model using a joint set of source and target images~\cite{he2022secret, li2022reliability, dai2021idm, zheng2021online, isobe2021towards, zheng2021group, ge2020mutual, zhao2020unsupervised, zheng2021exploiting, ge2020self}. We incorporate the CD loss for target images to consider the diversity of camera labels within each cluster. At test time, we compute L2 distances between query and gallery person representations to perform cross-camera matching. Note that the camera labels are used during training only.

\subsection{CaCL}
\label{subsec:curriculum_learning}
Given a target dataset, obtained from $C$ different cameras, we first divide the target dataset into multiple subsets by exploiting camera labels of person images. Concretely, we denote by $\mathcal{D}^{\textrm{S}}$ and $\mathcal{D}^{\textrm{T}}$ sets of images in the source and target domain datasets, respectively. We divide $\mathcal{D}^{\textrm{T}}$ into total $C$ number of non-overlapping subsets w.r.t camera labels, where each subset is denoted by $\mathcal{D}^{\textrm{T}}_{c}$, and $\mathcal{D}^{\textrm{T}} = \mathcal{D}^{\textrm{T}}_{1}\cup \mathcal{D}^{\textrm{T}}_{2}\cup \cdots \cup \mathcal{D}^{\textrm{T}}_{C}$. We start training a model using a single subset in the first curriculum stage, and incrementally expand the training set to $c$ subsets in the $c$-th stage for $c=1,\dots,C$, according to a curriculum sequence, computed by a camera-driven scheduling rule. That is, the curriculum sequence determines which subsets are used to increment the training set at each stage~(Fig.~\ref{fig:curriculum}). At each curriculum stage, we employ a self-training scheme~\cite{fan2018unsupervised} that alternates between clustering and fine-tuning steps.

\noindent \textbf{Camera-driven scheduler~(Fig.~\ref{fig:curriculum}).} Setting an effective training sequence plays an important role in curriculum learning~\cite{soviany2022curriculum, wang2021survey}. In the context of our approach to leveraging camera labels, the scheduling is equivalent to determining which camera in the target domain is easier to learn for a reID model trained on the source domain. We assume that knowledge transfer between domains of similar distributions is typically easier than the opposite case. We implement this idea using the MMD~\cite{gretton2006kernel} that computes distributional discrepancies between different domains. Specifically, we compute pairwise MMDs between the source dataset, $\mathcal{D}^{\textrm{S}}$, and target subsets, $\mathcal{D}^{\textrm{T}}_{c}$, by mapping the samples to the reproducing kernel Hilbert space $\mathcal{H}$ with a function $\phi(\cdot)$ associated with Gaussian kernel, as follows: 
\begin{equation}
   \label{eq:mmd}
   \text{MMD}_{c} = 
   \Big\Vert \frac{1}{|\mathcal{D}^{\textrm{S}}|} \sum_{\mathbf{x}^{\textrm{S}}_i \in \mathcal{D}^{\textrm{S}}} \phi(\mathbf{x}^{\textrm{S}}_i) 
   - \frac{1}{|\mathcal{D}^{\textrm{T}}_{c}|} \sum_{\mathbf{x}^{\textrm{T}}_j \in \mathcal{D}^{T}_c} \phi(\mathbf{x}^{\textrm{T}}_j) \Big\Vert_{\mathcal{H}}^2, 
\end{equation}
where $\mathbf{x}^{\textrm{S}}_{i}$ and $\mathbf{x}^{\textrm{T}}_{j}$ denote the $i$-th and $j$-th sample in $\mathcal{D}^{\textrm{S}}$ and $\mathcal{D}^{\textrm{T}}_{c}$, respectively, and $\vert \cdot \vert$ counts the total number of samples within a set. We establish a curriculum sequence by sorting the pairwise MMDs in an ascending order, that is, the closer subset w.r.t the source domain in terms of MMD is exploited earlier.

CaCL with a camera-driven scheduler provides the following advantages for UDA reID: First, it allows a smooth adaptation from source to target domains in a progressive manner. CaCL leverages a subset within a target domain that depicts a similar distribution with the source domain in earlier training stages, facilitating a smooth adaptation, compared to previous approaches using domain-level regimes. Second, our model starts to learn from person images obtained using a single camera, typically showing a weaker extent of intra-class variations, then progressively expands to other images captured from multiple cameras. This gradual expansion from simple to diverse scenarios encourages our model to better handle the inter-camera variations.

\begin{figure}[t]
   \begin{center}
      \includegraphics[width=.98\linewidth]{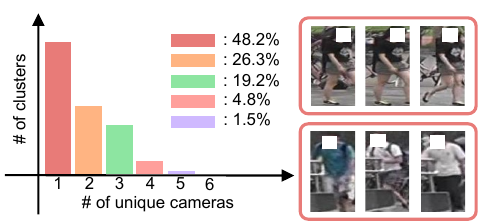}
      \vspace{-5mm}
   \end{center}
   \caption{\textbf{Left}:~Distribution of the number of clusters, across the number of unique cameras. We obtain the result on Market1501~\cite{zheng2015scalable} using a reID model trained on MSMT17~\cite{wei2018person}. Since the reID model trained on a single domain fails to generalize on other domains, most clusters simply contain person images captured by a single camera. \textbf{Right}:~Examples of person images within the same cluster. We can see that the person images do not show diverse intra-class variations~(top), and the clustering results are easily influenced by distracting cues~(\eg, occlusion in the left).}
   \label{fig:clustering}
\end{figure}
\begin{figure}[t]
   \begin{center}
      \includegraphics[width=.9\linewidth]{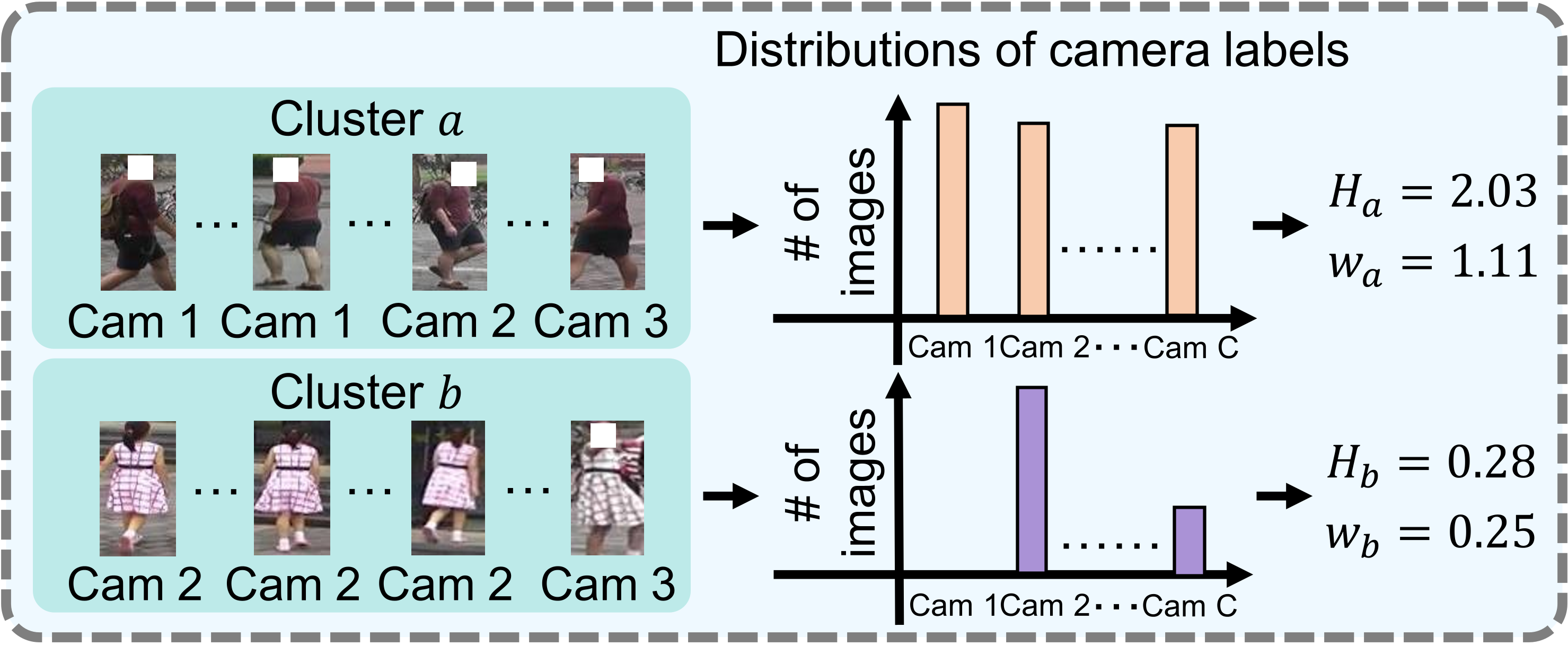}
      \vspace{-5mm}
   \end{center}
   \caption{An illustration of a detailed procedure for computing cluster-wise weighting factors. We compute the entropy of a camera distribution for each cluster, and then assign large weights for clusters with high entropy values. See text for details.}
   \vspace{-2mm}
   \label{fig:CDL}
\end{figure}

\subsection{CD loss}
\label{subsec:camera_diversity}

To generate pseudo ID labels for unlabeled target domain, previous methods~\cite{he2022secret, li2022reliability, dai2021idm, zheng2021online, isobe2021towards, zheng2021group, ge2020mutual, zhao2020unsupervised, zheng2021exploiting, ge2020self} apply a clustering algorithm on person representations of target images. However, we have observed that clustering results for target images are highly biased toward camera labels. We show in Fig.~\ref{fig:clustering} that most clusters~(up to $48.2\%$) are trivial ones that contain images obtained using a single camera. Directly training with the biased pseudo ID labels using standard cross-entropy~\cite{zheng2017discriminatively} and triplet~\cite{hermans2017defense} losses might be suboptimal, since the trivial clusters can dominate the adaptation process. Note that this is particularly important for reID that performs cross-camera image retrieval. To alleviate this issue, we propose to discard the trivial clusters, while encouraging the clusters of images from various cameras to involve more in feature learning. To this end, we measure the entropy of a camera distribution for each cluster, as follows:

\begin{equation}
   \label{eq:entropy}
   H_l = - \sum_{c} {r_{l}(c) \log (r_{l}(c))},
\end{equation}

\begin{algorithm}[t]
   \small
   \caption{Training}
   \begin{algorithmic}[1]
     \Require $N_c$: the number of iterations at $c$-th stage; $M_c$: an interval of generating pseudo labels at $c$-th stage; $\mathcal{A}$: an empty set; $C$: the number of cameras.
     \INPUT Source dataset $\mathcal{D}^{\textrm{S}}$; Target subsets $\mathcal{D}^{\textrm{T}}_{1}, \mathcal{D}^{\textrm{T}}_{2}, \dots, \mathcal{D}^{\textrm{T}}_{C}$.
     \OUTPUT A trained reID network.
     \State Pre-train a network using $\mathcal{D}^{\textrm{S}}$.
     \State Compute MMD between $\mathcal{D}^{\textrm{S}}$ and $\mathcal{D}^{\textrm{T}}_{c}$~[Eq.~(\ref{eq:mmd})].
     \State Determine a curriculum sequence of target subsets and obtain a list of ordered subsets $\mathcal{O}$
     \For {$c=1$ to $C$}
        \State $\mathcal{A}$ $\leftarrow$ $\mathcal{A}$ $\cup$ $\mathcal{O}(c)$
        \While {$i$ $\leq$ $N_{c}$}
           \If {($i$ mod $M_c$) $=$ $0$}
              \State Cluster images of $\mathcal{A}$ and generate pseudo labels.
              \State Measure cluster-wise camera entropy $H_l$~[Eq.~(\ref{eq:entropy})].
              \State Obtain cluster-wise weighting factor $w_l$~[Eq.~(\ref{eq:weight})].
           \EndIf
           \State Sample a mini-batch from $\mathcal{D}^{\textrm{S}}$ and $\mathcal{A}$.
           \If {$c = 1$}
              \State Update the network using $\mathcal{L}$ without $w_l$~[Eq.~(\ref{eq:loss})].
           \Else
              \State Update the network using $\mathcal{L}$ with $w_l$~[Eq.~(\ref{eq:loss})].
           \EndIf
        \EndWhile
     \EndFor
   \end{algorithmic}
   \label{alg:ours}
  \end{algorithm}
where $r_{l}(c) = \frac{n_{l}(c)}{\sum_c n_{l}(c)}$, and $n_{l}(c)$ is the number of images captured by the $c$-th camera within the $l$-th cluster. With the entropy for each cluster, we define cluster-wise weighting factor:
\begin{equation}
   \label{eq:weight}
   w_l = \log(H_l + 1). 
\end{equation}
Namely, for clusters of images obtained from the same camera (\ie, $H_l=0$), the weighting factor becomes zero and discards the clusters for training. On the other hand, for clusters of images captured by different cameras, the weighting factor encourages images in the clusters to involve more for feature learning~(Fig.~\ref{fig:CDL}).

We incorporate cluster-wise weights, $w_{l}$, to enhance cross-entropy and triplet terms used for training reID models. Concretely, given a person image $\mathbf{x}^{\textrm{T}}_{i}$ assigned to the $l$-th cluster, we define the CD cross-entropy term as follows:
\begin{equation}
   \mathcal{L}^{\textrm{T}}_\textrm{CDC} = \mathbb{E} [- w_{l} \log (p(l| \mathbf{x}^{\textrm{T}}_{i}))],
\end{equation}
where $p(l| \mathbf{x}^{\textrm{T}}_{i})$ is a softmax probability of $\mathbf{x}^{\textrm{T}}_{i}$ being classified to the $l$-th pseudo ID label. The CD triplet term~$\mathcal{L}^{\textrm{T}}_\textrm{CDT}$ is defined similarly. Note that for the first curriculum sequence, where input target images are taken from a single camera, we omit the weighting factors and employ vanilla cross-entropy and triplet losses for training.

\begin{table*}[!t]
\small
   \caption{Quantitative comparisons with the state of the art on a real-to-real scenario. Numbers in bold indicate the best performance and underscored ones indicate the second best. Results in parentheses are obtained with the source codes provided by the authors.}
   \label{tab:sota_realtoreal}
   \centering
   \begin{tabular}{l|c|cccc|cccc} 
      \hline
      \multirow{2}{*}{Methods}         & \multirow{2}{*}{Reference}     & \multicolumn{4}{c|}{MSMT17-to-Market1501}     & \multicolumn{4}{c}{Market1501-to-MSMT17}                       \\ \cline{3-10}
                                       &                & mAP           & R1            & R5            & R10           & mAP           & R1            & R5            & R10            \\ \hline
      MMT~\cite{ge2020mutual}          & ICLR 2020      & 75.6          & 89.3          & 95.8          & 97.5          & 22.9          & 49.2          & 63.1          & 68.8           \\
      SpCL~\cite{ge2020self}           & NeurIPS 2020   & 77.5          & 89.7          & 96.1          & 97.6          & 26.8          & 53.7          & 65.0          & 69.8           \\
      UNRN~\cite{zheng2021exploiting}  & AAAI 2021      & (78.3)        & (90.4)        & (96.5)        & (97.9)        & 25.3          & 52.4          & 64.7          & 69.7           \\
      GLT~\cite{zheng2021group}        & CVPR 2021      & (79.3)        & (90.7)        & (96.5)        & (98.0)        & 26.5          & 56.6          & 67.5          & 72.0           \\
      HCD~\cite{zheng2021online}       & ICCV 2021      & 80.2          & 91.4          & -             & -             & 28.4          & 54.9          & -             & -              \\
      IDM~\cite{dai2021idm}            & ICCV 2021      & \uline{82.1}  & \uline{92.4}  & \uline{97.5}  & \uline{98.4}  & 33.5          & 61.3          & 73.9          & 78.4           \\
      RESL~\cite{li2022reliability}    & AAAI 2022      & -             & -             & -             & -             & \uline{33.6}  & \uline{64.8}  & \uline{74.6}  & \uline{79.6}   \\
      \hline
      Ours                             &                & \textbf{84.7} & \textbf{93.8} & \textbf{97.7} & \textbf{98.6} & \textbf{36.5} & \textbf{66.6} & \textbf{75.3} & \textbf{80.1}  \\ \hline
      \end{tabular}
\end{table*}

\begin{table*}[!t]
\small
   \caption{Quantitative comparisons with the state of the art on a synthetic-to-real scenario. Numbers in bold indicate the best performance and underscored ones indicate the second best.}
   \label{tab:sota_syntoreal}
   \centering
   \begin{tabular}{l|c|cccc|cccc} 
      \hline
      \multirow{2}{*}{Methods} & \multirow{2}{*}{Reference} & \multicolumn{4}{c|}{PersonX-to-Market1501}                    & \multicolumn{4}{c}{PersonX-to-MSMT17}                          \\ 
      \cline{3-10}
                               &                            & mAP           & R1            & R5            & R10           & mAP           & R1            & R5            & R10            \\ 
      \hline
      MMT~\cite{ge2020mutual}  & ICLR 2020                  & 71.0          & 86.5          & 94.8          & 97.0          & 17.7          & 39.1          & 52.6          & 58.5           \\
      SpCL~\cite{ge2020self}   & NeurIPS 2020               & 73.8          & 88.0          & 95.3          & 96.9          & 22.7          & 47.7          & 60.0          & 65.5           \\
      IDM~\cite{dai2021idm}    & ICCV 2021                  & \uline{81.3}  & \uline{92.0}  & \uline{97.4}  & \uline{98.2}  & \uline{30.3}  & \uline{58.4}  & \uline{70.7}  & \uline{75.5}   \\ 
      \hline
      Ours                     &                            & \textbf{82.3} & \textbf{92.8} & \textbf{97.6} & \textbf{98.6} & \textbf{36.2} & \textbf{66.9} & \textbf{69.4} & \textbf{80.9}  \\ 
      \hline\hline
      \multirow{2}{*}{Methods} & \multirow{2}{*}{Reference} & \multicolumn{4}{c|}{Unreal-to-Market1501}                     & \multicolumn{4}{c}{Unreal-to-MSMT17}                           \\ 
      \cline{3-10}
                               &                            & mAP           & R1            & R5            & R10           & mAP           & R1            & R5            & R10            \\ 
      \hline
      JVTC~\cite{li2020joint}  & ECCV 2020                  & 78.3          & 90.8          & -             & -             & 25.0          & 53.7          & -             & -              \\
      IDM~\cite{dai2021idm}    & ICCV 2021                  & \uline{83.2}  & \uline{92.8}  & \uline{97.3}  & \uline{98.2}  & \uline{38.3}  & \uline{67.3}  & \uline{78.4}  & \uline{82.6}   \\ 
      \hline
      Ours                     &                            & \textbf{84.0} & \textbf{93.3} & \textbf{97.6} & \textbf{98.5} & \textbf{40.3} & \textbf{70.0} & \textbf{80.5} & \textbf{84.0}  \\
      \hline
      \end{tabular}
\end{table*}


\subsection{Overall training}
\label{subsec:overall_training}

We pre-train a reID model using ground-truth ID labels of source images using conventional cross-entropy~\cite{zheng2017discriminatively} and triplet~\cite{hermans2017defense} losses. We establish a curriculum with a camera-driven scheduler, and then perform clustering to obtain pseudo ID labels for target images. During fine-tuning, we exploit both source and target domains jointly, following~\cite{zheng2021exploiting, zheng2021online, isobe2021towards, zheng2021group, ge2020mutual, dai2021idm}. We adopt the cross-entropy loss~($\mathcal{L}^{\textrm{S}}_\textrm{CE}$) for source images, and the CD term~($\mathcal{L}^{\textrm{T}}_\textrm{CD}$) for target images, where the CD loss consists of CD cross-entropy and CD triplet terms. At each fine-tuning stage, we optimize a reID network with the overall objective as follows: 
\begin{equation}
   \label{eq:loss}
   \mathcal{L} =  \mathcal{L}^{\textrm{S}}_\textrm{CE} + \mathcal{L}^{\textrm{T}}_\textrm{CD}.
\end{equation}
We summarize in Algorithm~\ref{alg:ours} an overall training process of our approach.

\section{Experiments}

\subsection{Implementation details}
\label{subsec:imple}

\noindent \textbf{Dataset and evaluation metric.} We use four person reID datasets in our experiments, including Market1501~\cite{zheng2015scalable}, MSMT17~\cite{wei2018person}, PersonX~\cite{sun2019dissecting} and Unreal~\cite{zhang2021unrealperson}, where PersonX and Unreal provide synthetic images and corresponding ID labels. Market1501 contains pedestrian images of 1,501 IDs, captured by 6 cameras, where it consists of 12,936 images of 751 IDs for training and 19,732 images of 750 IDs for testing. MSMT17 contains 126,441 images, obtained from 15 cameras, where it consists of 32,621 images of 1,041 IDs and 93,820 images of 6,120 IDs for training and testing, respectively. PersonX and Unreal provide 9,840 and 130,244 images, respectively, for training. Following the evaluation protocol in UDA reID~\cite{dai2021idm, ge2020self, li2022reliability, zheng2021exploiting, zheng2021group}, we apply our approach to real-to-real and synthetic-to-real scenarios. We report the mean average precision~(mAP) and cumulative matching characteristics~(CMC) at rank-1, rank-5, and rank-10 for evaluation. 

\noindent \textbf{Training.} We adopt ResNet-50~\cite{he2016deep}, pre-trained for ImageNet classification~\cite{deng2009imagenet}, as a backbone network, where we use domain-specific BNs~\cite{chang2019domain} following ~\cite{dai2021idm,zheng2021group,zheng2021exploiting}. We train ResNet-50 with a source dataset, and use it as an initial reID model for UDA reID. We train the model for 4 epochs for each curriculum stage, except for the final stage, where we use 30 epochs, with the learning rate of $3.5 \times 10^{-4}$. Following ~\cite{zheng2021exploiting,zheng2021group,dai2021idm}, we set the batch size to 128, with 64 images from each domain. We use the Adam optimizer~\cite{kingma2014adam} with $\beta_{1}=0.9$ and $\beta_{2}=0.999$, and employ the XBM technique~\cite{wang2020cross} for triplet losses throughout all experiments, as done in~\cite{dai2021idm}. We use the DBSCAN~\cite{ester1996density} algorithm to cluster target images and generate pseudo ID labels, where we update the pseudo labels at every 3 epochs. Following ~\cite{zheng2021exploiting,zheng2021group,dai2021idm}, we resize the person image to the size of $256$ $\times$ $128$ and apply data augmentation techniques, including random flipping, random cropping, and random erasing~\cite{zhong2020random}. Detailed descriptions for hyperparameter settings are available in the supplement.  

\begin{table}[t]
\small
\SetTblrInner{rowsep=0.2pt, colsep=5pt}
\caption{Quantitative comparisons of variants of our model on Market1501~\cite{zheng2015scalable}-to-MSMT17~\cite{wei2018person} and MSMT17-to-Market1501. Numbers in bold indicate the best performance and underscored ones indicate the second best. M: Market1501, MS: MSMT17, RS: Random sequence, CTL: Cross-entropy~\cite{zheng2017discriminatively} and triplet~\cite{hermans2017defense} losses, CDL: CD loss.}
   \centering
   \label{Tab:ablation}
   \begin{tblr}{
      cells = {c},
      cell{1}{1} = {c=2}{},
      cell{1}{3} = {c=2}{},
      cell{1}{5} = {c=2}{},
      cell{1}{7} = {c=2}{},
      vline{3,5,7} = {-}{},
      hline{1-3,9} = {-}{},
    }
    Curriculum &            & Loss           &              & M-to-MS       &               & MS-to-M       &               \\
    RS         & CaCL       & CTL            & CDL          & mAP           & rank-1        & mAP           & rank-1        \\
               &            & \checkmark     &              & 23.4          & 50.4          & 79.2          & 92.3          \\
   \checkmark  &            & \checkmark     &              & 24.4          & 51.3          & 79.4          & 92.4          \\
   \checkmark  &            &                & \checkmark   & 30.4          & 58.5          & 81.4          & 92.8          \\
               & \checkmark & \checkmark     &              & \uline{31.1}  & \uline{59.9}  & 81.2          & 92.7          \\
               &            &                & \checkmark   & 30.1          & 58.3          & \uline{81.8}  & \uline{93.0}  \\
               & \checkmark &                & \checkmark   & \textbf{36.5} & \textbf{66.6} & \textbf{84.7} & \textbf{93.8} 
    \end{tblr}
\end{table}

\begin{table}[t]
\small
\SetTblrInner{rowsep=0.2pt}
   \caption{Quantitative comparisons between CD loss and UGID~\cite{zheng2021exploiting} loss on real-to-real and synthetic-to-real scenarios. Numbers in bold indicate the best performance and underscored ones indicate the second best. M: Market1501, MS: MSMT17, PX: PersonX, U: Unreal, CDL: Camera-diversity loss.}
   \centering
   \label{Tab:pseudolabel}
   \begin{tblr}{
      cells = {c},
      cell{1}{1} = {r=2}{},
      cell{1}{2} = {c=2}{},
      cell{1}{4} = {c=2}{},
      cell{6}{1} = {r=2}{},
      cell{6}{2} = {c=2}{},
      cell{6}{4} = {c=2}{},
      column{1} = {l},
      cell{1}{1} = {c},
      cell{6}{1} = {c},
      vline{2,4} = {1-6}{},
      vline{2,4} = {7-11}{},
      hline{1,3,6,8,11} = {-}{},
      hline{2} = {2-5}{},
      hline{7} = {2-5}{},
    }
    Variants                              & M-to-MS         &                 & MS-to-M         &        \\
                                          & mAP             & rank-1          & mAP             & rank-1 \\
    UGID~\cite{zheng2021exploiting}       & 32.3            & 61.0            & 82.2            & 92.8   \\
    CDL                                   & \uline{36.5}    & \uline{66.6}    & \uline{84.7}    & \uline{93.8}   \\
    CDL+UGID~\cite{zheng2021exploiting}   & \textbf{38.4}   & \textbf{67.9}   & \textbf{85.2}   & \textbf{94.1}  \\ \hline
    Variants                              & PX-to-M         &                 & U-to-MS         &        \\ 
                                          & mAP             & rank-1          & mAP             & rank-1 \\
    UGID~\cite{zheng2021exploiting}       & 81.2    		   & 91.9   		   & 35.2    		   & 63.4           \\
    CDL                                   &\uline{82.3}     &\uline{92.8}     &\uline{40.3}     &\uline{70.0} \\
    CDL+UGID~\cite{zheng2021exploiting}   &\textbf{83.1}    &\textbf{93.2}    &\textbf{41.2}    &\textbf{71.4}
    \end{tblr}
    \vspace{-1mm}
\end{table}

\subsection{Comparison with the state of the art}
\label{subsec:sota}

We compare our method with the state of the art on the real-to-real scenario in Table~\ref{tab:sota_realtoreal}. Overall, we can see from the results that our approach outperforms other methods on all benchmarks. UNRN~\cite{zheng2021exploiting} focuses on leveraging reliable labels, but does not consider the camera bias problem of pseudo labels. In contrast to UNRN, ours addresses the camera bias in pseudo labels, outperforming UNRN in all benchmarks by significant margins. IDM~\cite{dai2021idm} generates intermediate domains and leverages them to bridge source and target domains. However, it exploits all target images jointly during the adaptation process. In contrast to IDM, we address the large distribution gap of camera topologies between domains, by using camera labels of target images, outperforming IDM on all benchmarks. RESL~\cite{li2022reliability} also exploits camera labels of target images to train translation networks~\cite{zhu2017unpaired} that generate multiple images of the same person in the style of different cameras. On the contrary, we leverage camera labels of target images to establish a curriculum sequence and address the camera bias of pseudo labels. Our method outperforms RESL even without using the translation networks, indicating that our framework effectively leverages the camera labels to perform UDA reID. 

We provide in Table~\ref{tab:sota_syntoreal} a quantitative comparison between ours and state-of-art methods in the synthetic-to-real scenario. The results demonstrate that ours can effectively transfer the knowledge learned from the source domain to the target one, even for the synthetic-to-real scenario.  

\begin{figure*}[t]
   \begin{center}
   \includegraphics[width=.8\linewidth]{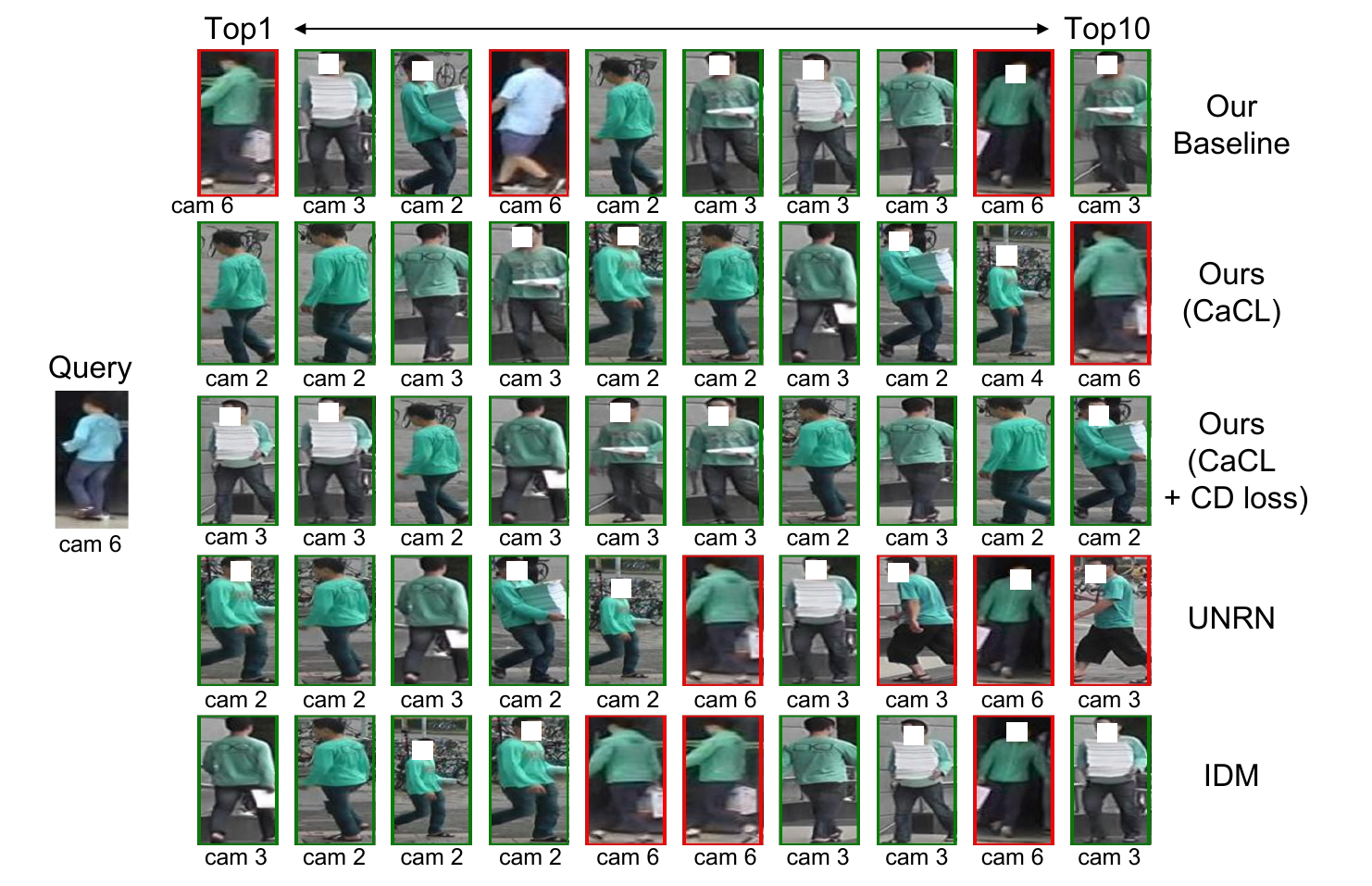}
   \end{center}
   \vspace{-4mm}
   \caption{Visual comparisons of retrieval results on MSMT17~\cite{wei2018person}-to-Market1501~\cite{zheng2015scalable}. Results with green boxes have the same identity as the query, while those with red boxes do not. (Best viewed in color.) }
   \vspace{-2mm}
   \label{fig:reID}
\end{figure*}

\subsection{Discussion}
\label{subsec:discussion}

\noindent \textbf{Ablation study.} We present in Table~\ref{Tab:ablation} an ablation analysis for each component of our method on Market1501-to-MSMT17 and MSMT17-to-Market1501. We report mAP and rank-1 scores for variants of our model. To validate the effectiveness of a camera-driven scheduler, we also provide results of setting a curriculum sequence randomly~(RS), and report the scores averaged over 5 trials. For the variants trained using cross-entropy and triplet losses~(CTL), we exclude the weighting factor within the CD loss for target images. We can see from the first and second rows that establishing a curriculum in a random sequence boosts the performance marginally, as this does not consider the discrepancies between target subsets and the source dataset. By exploiting the camera-driven scheduler in the fourth row, we boost the performance drastically. This coincides with findings reported in~\cite{soviany2022curriculum, wang2021survey}, that the result of incorporating curriculum depends on how the curriculum is designed. In this context, camera-driven scheduling rule provides a beneficial sequence for UDA reID, and enhances the adaptation performance. By comparing the fourth and the sixth rows, we can see that the CD loss further enhances the performance, confirming that incorporating cluster-wise weighting factors to selectively involve clusters is effective for adaptation. We can see from the first and fifth rows that CD loss still performs better than CTL even without a CaCL.

\noindent \textbf{Camera-diversity loss.} Similar to the weighting scheme in the CD loss, the work of~\cite{zheng2021exploiting} employs an UGID loss to re-weight the loss terms by measuring uncertainty among pseudo ID labels, and assigns large weights to the labels with low uncertainty values. For a comparison with the weighting scheme proposed in~\cite{zheng2021exploiting}, we show in Table~\ref{Tab:pseudolabel} results of models trained using the CD loss, the UGID-weighted loss, and a combination of the two. We simply multiply corresponding weight values from CD and UGID terms to exploit both losses. By comparing the first and the second rows, we can see that the model trained using the CD loss performs better than the one trained with the UGID for both cases. This suggests that addressing the inter-camera variations is more effective for UDA reID compared to handling samples with reliable labels. Note that computing uncertainty values for the UGID loss requires multiple reID models to measure prediction consistency, and thus demands additional computational complexity compared to the CD loss. We can also see that exploiting both losses in the third row shows the best performance, because the two weighting factors can complement each other. This suggests that combining our framework with other methods could lead to significant performance improvements in UDA reID.

\noindent \textbf{Qualitative analysis.} We show in Fig.~\ref{fig:reID} visual comparisons of retrieval results with the state of the art and variants of our model on MSMT17~\cite{wei2018person}-to-Market1501~\cite{zheng2015scalable}. The baseline is trained using the vanilla cross-entropy~\cite{zheng2017discriminatively} and triplet~\cite{hermans2017defense} losses without exploiting CaCL and the CD loss. We can see from the first and second rows that CaCL is more effective to retrieve person images, compared to the baseline. The third row shows that CaCL with the CD loss retrieves person images with the same IDs as the query correctly without the camera bias problem, confirming the effectiveness of the CD loss. Last three rows compare ours with other approaches (UNRN~\cite{zheng2021exploiting} and IDM~\cite{dai2021idm}). We can observe that they also retrieve person images with different IDs as the query, and suffer from the camera bias problem. In contrast, ours obtains accurate retrieval results, suggesting that it is robust to the camera bias problem of pseudo labels, effectively reducing inter-camera variations.

\section{Conclusion}
\vspace{-1mm}
We have presented a novel approach for UDA reID that performs a progressive adaptation by leveraging camera labels of person images. We propose a CaCL framework, gradually transferring the knowledge learned from a source domain to a target one, while addressing the large distribution gap of camera topologies between domains. We have also introduced a novel CD loss, mitigating a camera bias in pseudo labels and handling inter-camera variations, while progressively adapting a reID model from source to target domains. Experimental results show the effectiveness of our framework, setting a new state of the art on standard benchmarks. 

\vspace{3mm}
\noindent {\textbf{Acknowledgments.} 
This work was partly supported by the IITP and NRF grants funded by the Korea government(MSIT) (No.RS-2022-00143524, Development of Fundamental Technology and Integrated Solution for Next-Generation Automatic Artificial Intelligence System, No. 2023R1A2C2004306), and the Yonsei Signature Research Cluster Program of 2023 (2023-22-0008).
\clearpage

\newpage

{\small
\bibliographystyle{ieee_fullname}
\bibliography{CaCL}
}

\clearpage


\section*{Supplementary material}

# Camera-Driven Representation Learning for Unsupervised Domain Adaptive Person Re-identification - Supplementary materials

Geon Lee[1]  Sanghoon Lee[1]  Dohyung Kim[1]  Younghoon Shin[2]  Yongsang Yoon [2]  Bumsub Ham[1]*

[1]Yonsei University    [2]Robotics Lab, Hyundai Motor Company

In the supplementary material, we present training cost comparisons (Sec. S1), training epochs (Sec. S2), results for IBN-ResNet50 (Sec. S3), qualitative results (Sec. S4), and detailed descriptions for setting hyperparameters (Sec. S5).

## S1. Training cost comparisons

To demonstrate the efficiency of our approach, we compare the training cost with the state of the art [1, 3]. We measure GPU memory consumption during training and total training hours on MSMT17 [2]-to-Market1501 [4], and present the results in Table S1. For a fair comparison, we average the numbers over 10 executions using the official implementations provided by the authors[1], on the same machine with 4 Geforce RTX 2080Ti GPUs. Our approach offers faster training while using less GPU memory, because it does not employ multiple reID models [3], or explicitly generate more features [1], while outperforming them for all cases (Tables 1 and 2). Note that IDM even updates pseudo labels more frequently to maximize the performance.

## S2. Training epochs

Our model is trained with 4 epochs for each curriculum stage, except for the final one with 30 epochs. Namely, the total number of epochs is $4(C-1) + 30$, where $C$ is the number of cameras in a target domain. We use a half of target images on average for each stage, except for the final one with all images. The number of iterations is thus represented as $(2C - 2)\lceil \frac{N}{B} \rceil + 30 \lceil \frac{N}{B} \rceil$, where $N$ and $B$ are the number of target images and the batch size, respectively. The number of iterations for IDM [1] and UNRN [3] is $50 \lceil \frac{N}{B} \rceil$, as they use 50 epochs with all target images for entire stages. If $C$ is smaller than 11, e.g., Market1501 [4]

---

*Corresponding author

[1]We directly adopt official codes from https://github.com/zkcys001/UDAStrongBaseline and https://github.com/SikaStar/IDM for UNRN [3] and IDM [1], respectively.

Table S1: Quantitative comparisons of ours, UNRN [3], and IDM [1] in terms of GPU memory consumption and training hours.

| Methods | GPU memory | Training hours |
|---|---|---|
| Ours | **27GB** | **3.2 hours** |
| UNRN [3] | 29GB | 4.3 hours |
| IDM [1] | 54GB | 6.8 hours |

captured by 6 cameras, ours uses fewer iterations than IDM and UNRN. It requires more iterations for MSMT17 [2] with 15 cameras.

## S3. Results for IBN-ResNet50

We have adopted IBN-ResNet50 as a backbone network, following IDM [1], and obtain the result on Market1501 [4]-to-MSMT17 [2]. Our model outperforms the IDM counterpart by 2.9% and 2.1% in terms of mAP and rank-1, respectively, demonstrating the effectiveness of our approach for UDA reID.

## S4. Qualitative results

We provide in Fig. S1 additional visual comparisons of retrieval results among variants of our model on MSMT17 [2]-to-Market1501 [4], and Market1501-to-MSMT17. We can see that the baseline model is distracted by different persons with, e.g., illumination (top-left), and similar pose (bottom-left). On the contrary, our model is able to offer person representations that are robust to various intra-class variations, and successfully retrieves person images with the same ID as a query person.

## S5. Hyperparameters

We mainly adopt hyperparameter settings from recent works [1, 3] (e.g., batch size, learning rate, and momen-

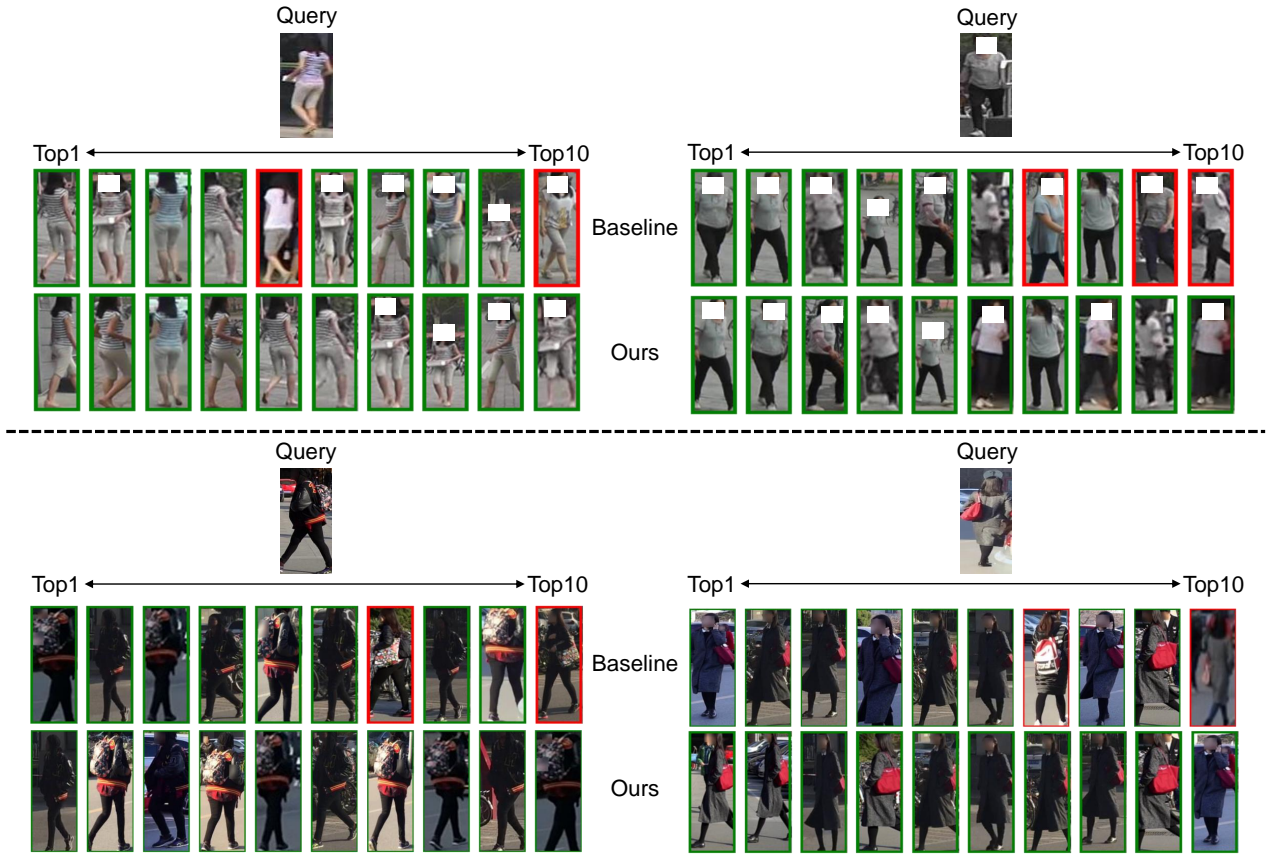

Figure S1: Visual comparisons of retrieval results on MSMT17 [2]-to-Market1501 [4] (top) and Market1501-to-MSMT17 (bottom). Results with green boxes have the same identity as the query, while those with red boxes do not. (Best viewed in color.)

Table S2: Quantitative comparisons for the hyperparameter $\epsilon_{init}$ and $M_{int}$. Numbers in bold indicate the best performance and underscored ones indicate the second best.

| $\epsilon_{init}$ | rank-1 (%) | mAP (%) |
|---|---|---|
| 0.3 | 70.2 ± 1.5 | 45.1 ± 1.0 |
| 0.4 | **71.1 ± 1.0** | **46.3 ± 0.5** |
| 0.5 | 70.1 ± 0.8 | 44.9 ± 1.1 |
| $M_{int}$ | rank-1 (%) | mAP (%) |
| 1 | 71.0 ± 0.4 | 46.1 ± 0.4 |
| 3 | **71.1 ± 0.5** | **46.3 ± 0.3** |
| 5 | 69.2 ± 1.0 | 44.5 ± 0.2 |

tum value for an EMA update), except for cluster density threshold $\epsilon$ for the DBSCAN algorithm and the frequency of updating pseudo labels $M_{int}$. We randomly split 1041 training IDs in MSMT17 [2] into 841 and 200 IDs for training and validation, respectively, and perform cross-validation on Market1501 [4]-to-MSMT17 [2]. We set the threshold value to $\epsilon_{init}$ during the initial curriculum stage, and linearly increase the value at each stage, up to 0.6 at the final stage. This is consistent with recent works [1, 3] that set $\epsilon$ to 0.6 for training with all target images. We perform a grid search for the initial density threshold $\epsilon_{init}$ over values in $\{0.2, 0.3, 0.4, 0.5\}$. We perform a grid search for $M_{int}$ over $\{1, 3, 5\}$. We adopt the corresponding $\epsilon_{init}$ and $M_{int}$ values across all settings and scenarios. We provide in Table S2 the results for various $\epsilon_{init}$ and $M_{int}$ values.



\end{document}